\pdfoutput=1
\documentclass[11pt]{article}
\usepackage{graphicx}
\graphicspath{ {./images/} }

\usepackage{acl}
\usepackage{times}
\usepackage{latexsym}
\usepackage{xcolor}

\usepackage[T1]{fontenc}
\usepackage[utf8]{inputenc}

\usepackage{microtype}

\usepackage{booktabs}
\usepackage{listings}

\usepackage{textgreek}
\usepackage{dirtytalk}
\usepackage{caption}

\begin{document}

% If the title and author information does not fit in the area allocated, uncomment the following
%
%\setlength\titlebox{<dim>}
%
% and set <dim> to something 5cm or larger.
\title{Automatic Fake News Detection: Are current models ``fact-checking'' or ``gut-checking''?}

\author{Ian Kelk\hspace{1cm}Benjamin Basseri\hspace{1cm}Wee Yi Lee\hspace{1cm}Richard Qiu\hspace{1cm}Chris Tanner\\ \\
        Harvard University \\ \texttt{\{iak415@g,basseri@cs50,wel390@g\}.harvard.edu}\\
        \texttt{\{rqiu@college,christanner@g\}.harvard.edu}}
% \author{Ian Kelk \And Benjamin Basseri \And Wee Yi Lee \And Richard Qiu \And Chris Tanner} 
% Harvard University\\
% % \texttt{\{iak415@g,basseri@cs50,wel390@g,rqiu@college,christanner@g\}.harvard.edu\}}

\maketitle

\begin{abstract}
Automatic fake news detection models are ostensibly based on logic, where the truth of a claim made in a headline can be determined by supporting or refuting evidence found in a resulting web query. These models are believed to be reasoning in some way; however, it has been shown that these same results, or better, can be achieved without considering the claim at all – only the evidence. This implies that other signals are contained within the examined evidence, and could be based on manipulable factors such as emotion, sentiment, or part-of-speech (POS) frequencies, which are vulnerable to adversarial inputs. We neutralize some of these signals through multiple forms of both neural and non-neural pre-processing and style transfer, and find that this flattening of extraneous indicators can induce the models to actually require both claims and evidence to perform well.  We conclude with the construction of a model using emotion vectors built off a lexicon and passed through an ``emotional attention'' mechanism to appropriately weight certain emotions. We provide quantifiable results that prove our hypothesis that manipulable features are being used for fact-checking. 
\end{abstract}

\section{Introduction}
Recent events such as the last two U.S. presidential elections have been greatly affected by fake news, defined as ``fabricated information that disseminates deceptive content, or grossly distort actual news reports, shared on social media platforms'' \citep{10.1257/jep.31.2.211}. In fact, the World Economic Forum 2013 report designates massive digital misinformation as a major technological and geopolitical risk \citep{wildfire}. As daily social media usage increases \href{https://www.statista.com/topics/7863/social-media-use-during-coronavirus-covid-19-worldwide}{(Statista Research Department, 2021)}, manual fact-checking cannot keep up with this deluge of information.

\begin{figure*}[t!]
\centering
  \caption{An example of why evidence alone does not suffice in identifying fake news, despite the evidence being conditioned on the claim as a search-engine query. Although the returned evidence appearing reputable, it is clear that it has little relevance to deciding the veracity of the claim that "all Canadians have eaten at least one bear."}
  \includegraphics[width=1.0\textwidth]{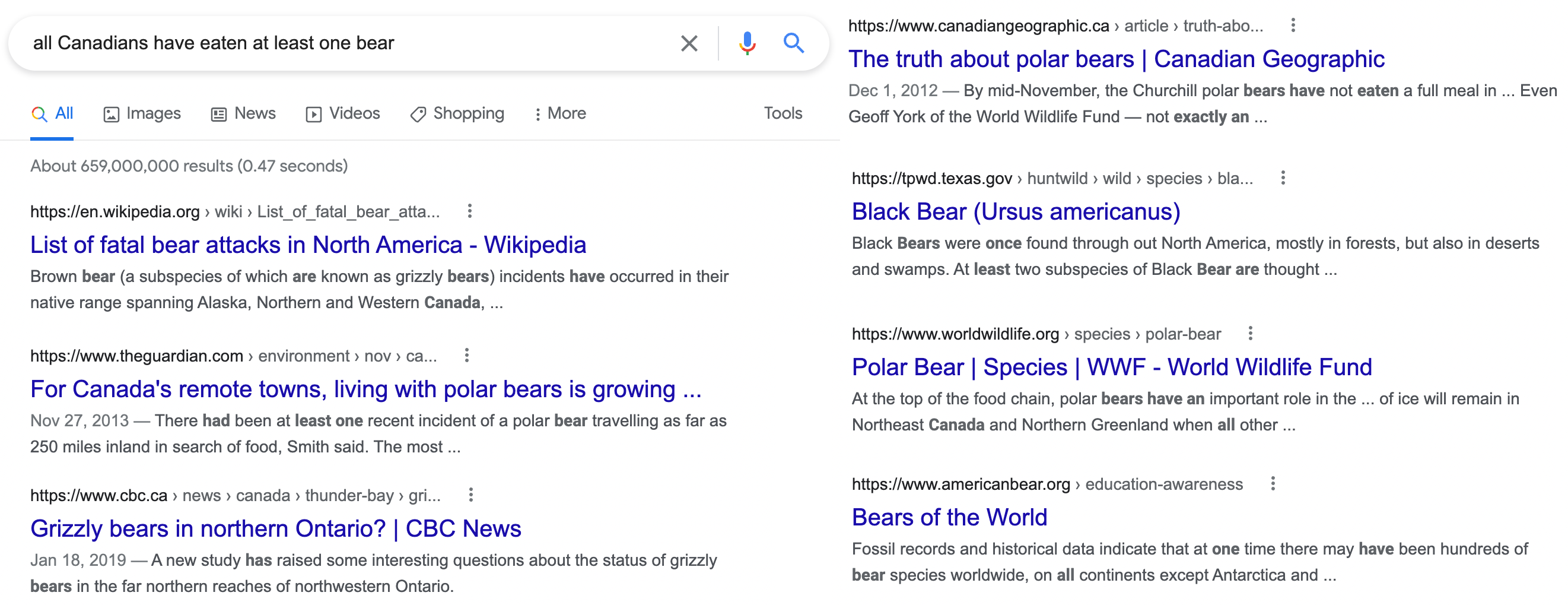}
  \label{fig:canadian-bear}
\end{figure*}

Automatic fact-checking models are therefore a necessity, and most of them function using a system of \textit{claims} and \textit{evidence} \citep{10.1145/3097983.3098131}. Given a specific claim, the models use external knowledge as evidence. Typically, a web search query is treated as the claim, and a subset of the top search results is treated as the evidence. There is an implicit assumption that the fact-checking models are reasoning in some way, using the evidence to confirm or refute the claim. Recent research \citep{hansen2021automatic} found this conclusion may be premature; current models can show improved performance when considering evidence alone, essentially fact-checking an unasked question.
% EDIT START
While this might seem reasonable given that the evidence is conditioned on the claims by the search engine, this can be exploited as illustrated in Figure ~\ref{fig:canadian-bear}, which shows that evidence returned using a ridiculous claim can still appear reasonable if we view the evidence alone without the claim.
Furthermore, textual entailment requires both a text and a hypothesis; if we have a result without a hypothesis, we are performing a different, unknown task.

% EDIT END

This finding indicates a problem with current automatic fake news detection, signaling that the models rely on features in the evidence typical to fake news, rather than using entailment. Since most automated fact-checking research is primarily concerned with the accuracy of the results, rather than addressing \textit{how} the results are achieved, we propose a novel investigation into these models and their evidence. We use a variety of pre-processing steps, including neural and non-neural ones, to attempt to reduce the affectations common in evidence:

\begin{itemize}
\item Stemming, stopword removal, negation, and POS-filtering  \citep{babanejad-etal-2020-comprehensive}.
\item Style transfer neural models using the \href{https://github.com/PrithivirajDamodaran/Styleformer}{\textit{Styleformer}} model to perform \textbf{informal-to-formal} and \textbf{formal-to-informal} paraphrasing methods \citep{DBLP:journals/corr/abs-1804-06437, Schmidt2020GenerativeTS}.
\end{itemize}

We also develop our own BERT-based model as an extension of the \textit{EmoCred} system \citep{10.1145/3331184.3331285}, adding an ``emotional attention'' layer to weight the most relevant emotional signals in a given evidence snippet. We make our code publicly available. \footnote{\href{https://github.com/automatic-fake-news-detection}{GitHub repository link}}

With each of these methods, we focus on scores where the models perform better using \textbf{both the claims and the evidence combined}, $S_{C \& E}$, rather than with the \textbf{evidence alone}, $S_E$. Going forward, we will refer to the difference between these dataset combinations as the $delta$ of the pre-processing step, where $delta = S_{C\&E} - S_E$. A positive $delta$ score indicates that the claim was useful and helped yield an increase in performance. Since we are removing indicators that the current models rely on, some of the models perform \textit{worse} at the task than they did previously. However, a surprising result is that many \textit{improved}, and the need to consider the claim and the evidence together is a sign of using reasoning rather than manipulable indicators.

Under current fact-checking models, adversarial data can subvert these detectors. Paraphrasing can be performed by inserting fictitious statements into otherwise truthful evidence with little effect on the model's output. For example, an article titled ``Is the GOP losing Walmart?'', could have ``Walmart'' substituted with ``Apple,'' and the predictions are nearly identical despite the news now being fictitious \citep{DBLP:journals/corr/abs-1901-09657}.

\begin{figure*}[t!]
\centering
  \caption{Ablation studies where evidence was sequentially removed for training and evaluation of models. On the far left, we show the most effective non-neural pre-processing compared to the baseline of \textbf{none}. Performance generally worsens as the ablation increases.}
  \includegraphics[width=1.0\textwidth]{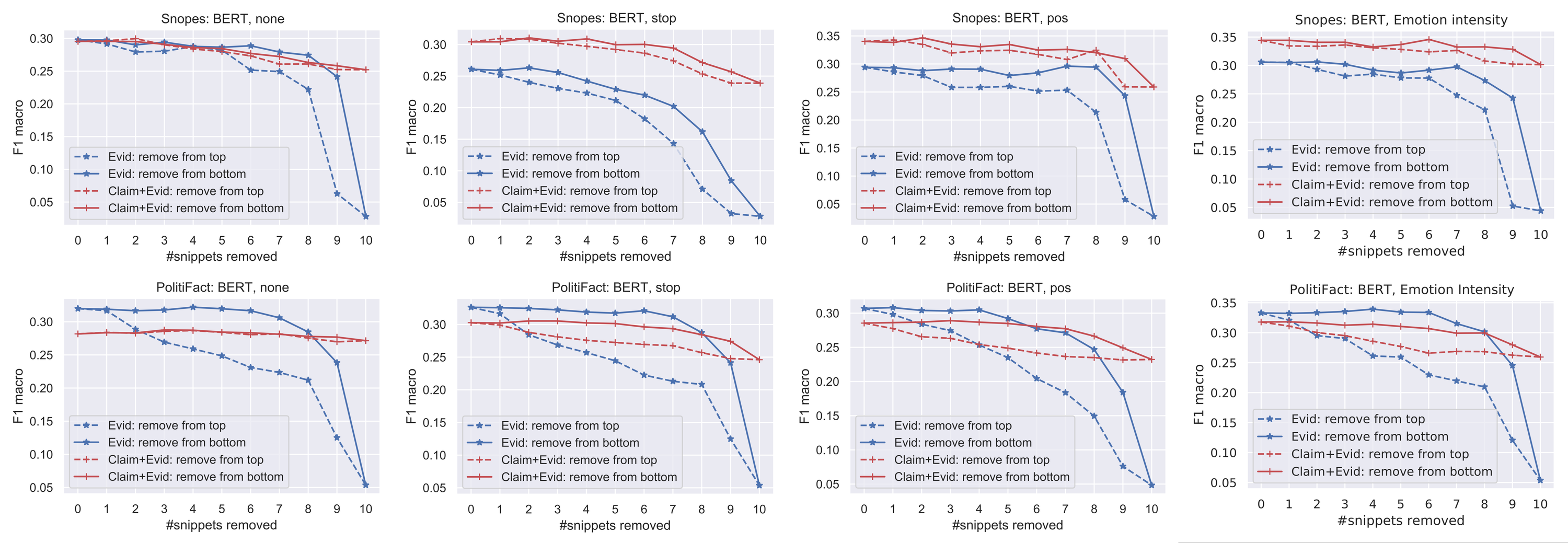}
  \label{fig:jack-output}
\end{figure*}

\section{Related Work}
There has been significant work with automatic fact-checking models using RNNs and Transformers \citep{DBLP:journals/corr/abs-2005-06058,DBLP:journals/corr/abs-2005-00033, Shaar2020OverviewOC} as well as non-neural machine learning using TF-IDF vectors \citep{DBLP:journals/corr/abs-1809-00509}.

Current fake news detection models that use a claim's search engine results as evidence may unintentionally use hidden signals that are not attributed to the claim \citep{hansen2021automatic}. Additionally, models may in fact simply memorize biases within data \citep{gururangan-etal-2018-annotation}. Improvements can be made when using human-identified justifications for fact-checking \citep{alhindi-etal-2018-evidence, vo-lee-2020-facts}, and making use of textual entailment can offer improvements \citep{inbook}. 

Emotional text can signal low credibility \citep{rashkin-etal-2017-truth}, characterizing fake news as a task where pre-processing can be used effectively to diminish bias \citep{10.1145/3331184.3331285, babanejad-etal-2020-comprehensive}. 
A framework to both categorize fake news and to identify features that differentiate fake news from real news has been described by \citet{doi:10.1177/0002764219878224}, and debiasing inappropriate subjectivity in text can be accomplished by replacing a single biased word in each sentence \citep{article}.

\section{Datasets}
We use the MultiFC dataset \citep{augenstein2019multifc}, which consists of political claims and associated truth labels from PolitiFact and Snopes. Using the claim as a query, the top ten results from Google News (``snippets'') constitute the evidence \citep{hansen2021automatic}. PolitiFact and Snopes use five labels (False, Mostly False, Mixture, Mostly True, True), which we collapse to True, Mixture, and False.

To construct the emotion vectors for our \textit{EmoAttention} system, we use the NRC Affect Intensity Lexicon, which maps approximately 6,000 terms to values between 0 and 1, representing the term's intensity along 8 different emotions \citep{DBLP:journals/corr/Mohammad17}. For example, ``interrupt'' and ``rage'' are both categorized as \textit{anger} words, but with the respective intensity values of 0.333 and 0.911.

\section{Models}
The most common automatic fact-checking NLP models are based on term frequency, word embeddings, and contextualized word embeddings, using Random Forests, LSTMs, and BERT \citep{10.1145/3097983.3098131}. We limit our experimentation to the BERT model, as it is the highest performing state-of-the-art model and was thoroughly tested in \citep{hansen2021automatic}. This BERT model with no pre-processing is our baseline model.

For the style transfer model we use the \href{https://github.com/PrithivirajDamodaran/Styleformer}{Styleformer model} \citep{DBLP:journals/corr/abs-1804-06437, Schmidt2020GenerativeTS}, a Transformer-based seq2seq model.

We also develop our own BERT-based model using the \textit{EmoLexi} and \textit{EmoInt} implementation of the EmoCred system by adding an \textit{emotional attention layer} to emphasize certain emotion representations for a given claim and its evidence \citep{10.1145/3331184.3331285}. There is also a \textit{snippet attention layer} attending to which evidence itself should be weighted most heavily for the given claim.

\begin{figure}[h!]
  \caption{The \textit{EmoAttention} BERT model architecture using \textit{emotional-} and \textit{snippet attention}}
  \includegraphics[width=0.5\textwidth]{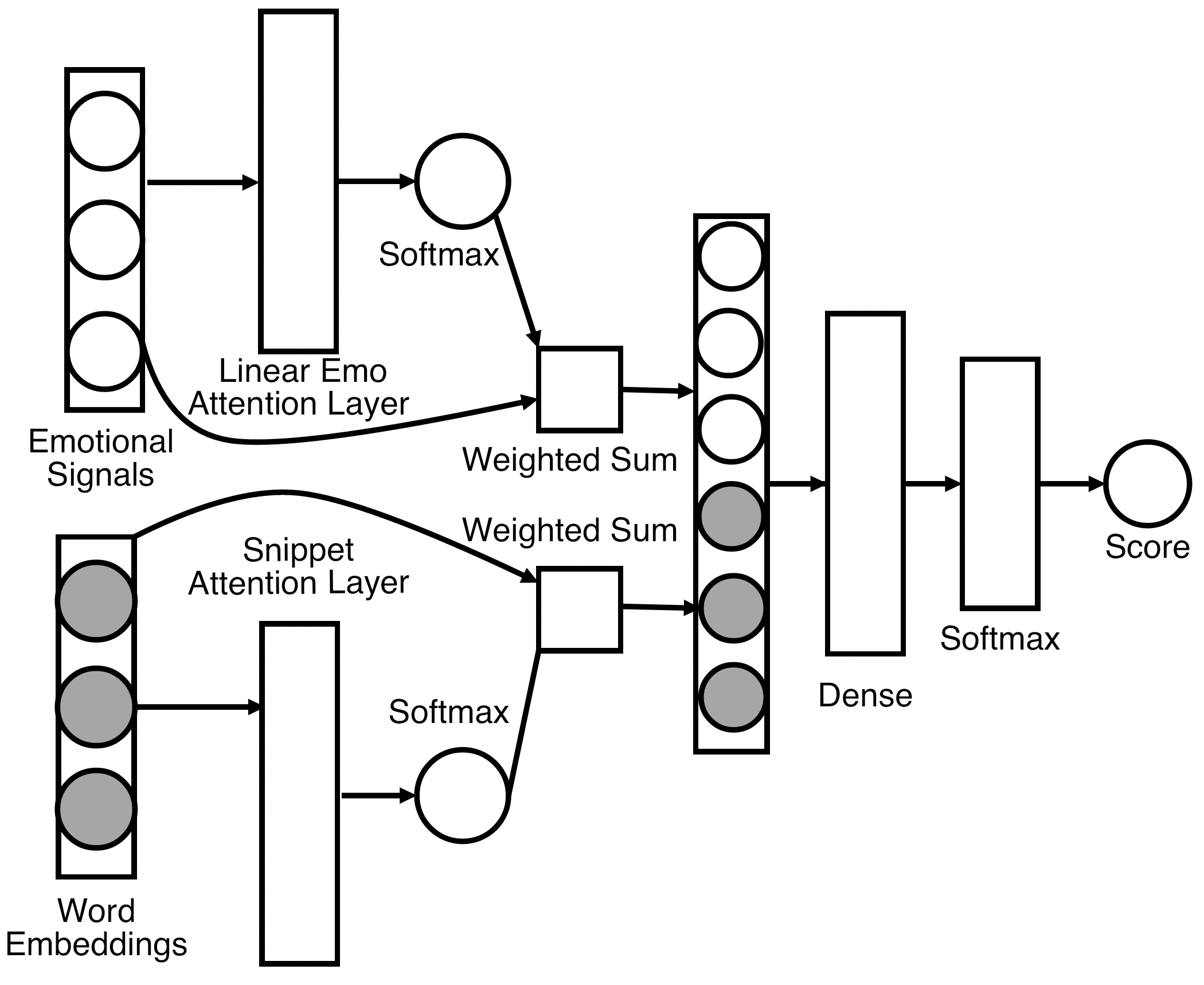}
  \label{fig:emocred-attention}
\end{figure}

\section{Experiments}

\subsection{Non-neural pre-processing}

% TANNER START
Our goal is to separate affect-based properties from factual content of the text. Toward this, we run a large number of permutations of the following four simple pre-processing steps (see Figure ~\ref{fig:full-chart} in Appendix B for results). These steps were chosen as they have been shown to facilitate affective tasks such as sentiment analysis, emotion classification, and sarcasm detection \citep{babanejad-etal-2020-comprehensive}. In some cases we used a modified form --- such as removing adverbs for POS pre-processing.
% TANNER END

%We run a large number of permutations of the following four simple pre-processing steps (see Figure ~\ref{fig:full-chart} in Appendix B for results).
% EDIT START
%These steps were chosen as they have been shown to facilitate affective tasks such as sentiment analysis, emotion classification, and sarcasm detection \citep{babanejad-etal-2020-comprehensive}. In some cases we used a modified form --- such as removing adverbs for POS pre-processing --- since the goal is to actually separate affect from the factual content of the text:
% EDIT END

\begin{itemize}
\item \textbf{Negation (NEG):} A mechanism that transforms a negated statement into its inverse \citep{10.5555/2392701.2392703}. An example, ``I am not happy'' would have ``not'' removed and ``happy'' replaced by its antonym, forming the sentence ``I am sad.''
\item \textbf{Parts-of-Speech (POS):} We keep only three parts of speech:  nouns, verbs, and adjectives. We initially included adverbs but found removing them improved results. This could be due to some adverbs being emotionally charged.
\item \textbf{Stopwords (STOP):} These are generally the most common words in a language, such as function words and prepositions. We use the NLTK library.
\item \textbf{Stemming (STEM):} Reducing a word to its root form. We use the NLTK Snowball Stemmer.
\end{itemize}

\begin{table*}
\centering
\begin{tabular}{lrrrr}
 \multicolumn{1}{c}{} & \multicolumn{2}{c}{Snopes} & \multicolumn{2}{c}{PolitiFact} \\
 \toprule
 Pre-processing  & $S_{C \& E}$  & \textDelta vs $S_{E}$ & $S_{C \& E}$ & \textDelta vs $S_{E}$ \\
  & (Claim+Evidence) & (Evidence) & (Claim+Evidence) & (Evidence) \\
  & F1 Macro & F1 Macro & F1 Macro & F1 Macro \\
 \midrule
 \hspace*{2mm}None & \color{red}0.295 & \color{red}-0.003 & \color{red}0.282 & \color{red}-0.038 \\
 \hspace*{2mm}POS & \color{blue}0.340 & \color{blue}0.046 & 0.285 & \color{blue}-0.022 \\
 \hspace*{2mm}STOP & 0.304 & \color{blue}0.043 & \color{blue}0.303 & -0.023 \\
 \hspace*{2mm}EmoAttention (EmoInt) & \color{blue}0.344 & \color{blue}0.038 & \color{blue}0.318 & \color{blue}-0.015 \\
 \hspace*{2mm}EmoAttention (EmoLexi) & 0.324 & \color{red}-0.003 & \color{blue}0.310 & -0.033 \\
 \hspace*{2mm}POS+STOP & 0.312 & 0.012 & 0.290 & \color{blue}-0.003 \\
 \hspace*{2mm}Formal to Informal & \color{blue}0.332 & 0.028 & –– & –– \\
 \bottomrule
\end{tabular}
\caption{\label{tab:Models} Top results from various pre-processing steps. The top three steps are highlighted in blue. The lowest F1 Macro scores and deltas are in red. With the exception of \textit{EmoLexi} tying for the lowest delta, the best pre-processing steps outperform the baseline BERT model from \citet{hansen2021automatic}.}
\label{table:BEN}
\end{table*}

\subsection{Neural formality style transfer}

We use the adversarial technique of generating paraphrases for all the claims and evidence through style transfer. The neural Transformer-based seq2seq model \textit{Styleformer} changes the formality of the text, and it frequently changes the ordering of the sentence itself, too. For example, the formal-to-informal model changes \textit{``A photograph shows William Harley and Arthur Davidson unveiling their first motorcycle in 1914''} to \textit{``In a 1914 photograph William Harley and Arthur Davidson unveil their first motorcycle.''}

As well, it removes punctuation and alters phrasing that might be understood as sarcasm, such as \textit{``Melania Trump said that Native Americans upset about the Dakota Access Pipeline should `go back to India'''} to \textit{``Melania Trump told Native Americans that was upset by the Dakota Access Pipeline, that they should travel to India.''} The informal-to-formal model lowercases everything and also changes the text significantly.

% EDIT START
We chose this paraphrasing model based on the idea that fake news -- especially that which is frequently posted on social media -- has a certain polarizing style that might be neutralized by altering the formality of the text. Rather surprisingly, we received better results transforming the style from formal-to-informal than we did with informal-to-formal.
% EDIT END
\subsection{EmoCred emotion representations with emotional attention}

The \textit{EmoCred} systems of \textit{EmoLexi} and \textit{EmoInt} use a lexicon to determine emotional word counts and intensities, respectively \citep{10.1145/3331184.3331285}.
% EDIT START
We use the \textit{NRC Affect Intensity Lexicon}, a ``high-coverage lexicons that captures word–affect intensities'' for eight basic emotions, which were created using a technique called best–worst scaling \citep{DBLP:journals/corr/Mohammad17}. These eight emotions can be used to create an emotion vector for a sentence, where each index corresponds to a score: [\textit{anger, anticipation, disgust, fear, joy, sadness, surprise, trust}].

As an example, a sentence that contains the word ``suffering'' conveys \textit{sadness} with an \textit{NRC Affect Intensity Lexicon} intensity of 0.844, whereas the word ``affection'' indicates \textit{joy} with an intensity of 0.647. We create the vector of length eight, and for each word associated with an emotion, the emotion's indexed value is either: (1) incremented by one for \textit{EmoLexi}; or, (2) incremented by its intensity for \textit{EmoInt}. Thus, the sentence ``He had an affection for suffering'' would have an \textit{EmoLexi} emotion vector of $[0,0,0,0,1,1,0,0]$ and an \textit{EmoInt} emotion vector of $[0,0,0,0,0.647,0.844,0,0]$
% EDIT END

We build on this \textit{EmoCred} framework, adding an attention system for emotion that gives a weight to each emotion vector, just as the attention layer for each snippet gives a weight to each snippet. The end result is that two independent attention layers attend to the ten snippets and ten emotional representations independently, and we call the resulting system \textit{Emotional Attention} (see Figure ~\ref{fig:emocred-attention}).

\section{Results}

Surprisingly, the four top-performing models with the Snopes dataset include two non-neural models and two neural models. All four achieve greater F1 Macro scores than the baseline BERT model without pre-processing (see Figure ~\ref{fig:jack-output}). POS and STOP yield the biggest delta between $S_{C \& E}$ vs. $S_{E}$, followed by \textit{EmoInt} and \textit{Informal Style Transfer}. However, \textit{EmoInt} yields the highest F1 Macro, followed by POS, \textit{Informal}, and STOP.

In PolitiFact, none of the pre-processing steps achieve a delta greater than zero for $S_{C \& E}$ versus $S_{E}$. The combination of POS+STOP steps comes closest to parity, followed by \textit{EmoInt}, then POS and STOP. For the best F1 Macro scores overall, \textit{EmoAttention}'s two forms (i.e., \textit{EmoInt} and \textit{EmoLexi}) were the two best, followed by STOP and POS. All of these pre-processing steps achieve higher F1 Macro scores than the baseline BERT model. Further, they yield better deltas for $S_{C \& E}$ versus $S_{E}$, implying that the model now requires the claims to reason.

\section{Conclusion}

Many pre-processing steps increase both the model's F1 scores and its need for claims \textit{and} evidence, validating our hypothesis that signals in style and tone have become a crutch for fact-checking models. Rather than doing entailment, they are leveraging other signals -- perhaps similar to sentiment analysis -- and relying on a ``gut feeling''. \textit{EmoAttention} generates our best predictions and deltas, confirming our suspicion that the models rely on emotionally charged style as a predictive feature. This is further narrowed to emotional \textit{intensity}: the \textit{EmoInt} intensity score-based model performs much better than its count-based counterpart \textit{EmoLexi}. Thus, evidence containing emotions associated with fake news will be considered more when scoring the claim.

One surprising result is the effectiveness of the simple POS and STOP pre-processing steps. POS only included nouns, verbs, and adjectives (i.e., a superset of STOP). This could explain why it has the best delta between $S_{C \& E}$ vs. $S_{E}$. Future research could investigate if stopwords, which are often discarded, actually contain signals such as anaphora: a repetitive rhetoric style which can affect NLP analyses \citep{LIDDY199039}.

As an example, Donald Trump makes heavy use of anaphora in his 2017 inauguration speech: 

\begin{quote}
\small{
\say{Together, \textbf{we will} make America strong \textbf{again}. \textbf{We will} make America wealthy \textbf{again}. \textbf{We will} make America proud \textbf{again}. \textbf{We will} make America safe \textbf{again}. And, yes, together, \textbf{we will} make america great \textbf{again}.} \href{https://trumpwhitehouse.archives.gov/briefings-statements/the-inaugural-address/}{(Trump Inauguration Address, 2017)}}
    
\end{quote}

By removing stopwords ``we'', ``will'' and ``again'', the model relies less on the text's rhetoric style and more on the entailment we are seeking. We propose further study on the effects of STOP and POS, as well as experimenting with different emotional vectors and \textit{EmoAttention} to make fact-checking models more robust. Automatic Fake News detection remains a challenging problem, and unfortunately, current fact-checking models can be subverted by adversarial techniques that exploit emotionally charged writing.

\appendix

\section{Impact Statement}

Disinformation is much more than just a mild inconvenience for society; it has resulted in needless deaths in the COVID-19 pandemic, and has fomented violence and political instability all over the globe \citep{10.3389/fpsyg.2020.566790}. Our goal in this paper is to discover exploitable weaknesses in current fact-checking models and recommend that such models not be relied upon in their current form. We point out how the models are dependent on emotional signals in the texts instead of exclusively performing textual entailment, and that additional research needs to be done to ensure they are performing the proper task.

\textbf{Harm Minimization} Our quantifying of the effects of pre-processing on fact-checking models does not cause any harm to real-world users or companies. Research has demonstrated that adversarial attacks could result in disinformation being labeled as factual news. Disinformation has become increasingly present in global politics, as some nation-states with significant resources have disseminated propaganda to create political dissent in other countries \citep{DBLP:journals/corr/abs-1901-09657}. Our research here has demonstrated potential risks: emotional writing could be used as an exploit to circumvent fact-checking models. Thus, we urge others to further illuminate such vulnerabilities, to minimize potential harms, and to encourage improvements with new models.

\textbf{Deployment} Social media companies often deal with fake news by placing highly visible labels. However, simply tagging stories as false can make readers more willing to believe and share \textit{other} false, untagged stories. This unintended consequence -- in which the selective labeling of false news makes other news stories seem more legitimate -- has been called the ``implied-truth effect'' \citep{author-implied-truth}. Thus, unless these models become so accurate that they catch \textit{all} fake news presented to them, the entire basis of their use is called into question.

Despite the significant progress in developing models to correctly identify fake news, the real elephant in the room is that many people simply ignore the labels \citep{doi:10.1177/0002764219878224}. There is, however, prior work supporting the idea that if people are warned that a headline is false, they will be less likely to believe it \citep{article-warnings, doi:10.1177/1529100612451018}. Because of this, we believe this research represents a net benefit for humanity.

Warning labels are just one way of dealing with properly identified fake news, and publishers can choose to simply not allow it on their platforms. Of course, this issue leads to questions of censorship.

% \clearpage
\section{Extended Results}

\begin{figure*}[t!]
\centering
  \caption{The full table of results for all pre-processing steps for the Snopes (SNES) and PolitiFact (POMT) datasets. Due to the high compute requirements of the formal and informal style transfer models, these datasets were only prepared for the Snopes dataset. The darkest green colors indicate the best results, while the red indicates the worst. Multiple pre-processing steps such as (pos, stop) were performed in the order written.}
  \includegraphics[width=1.0\textwidth]{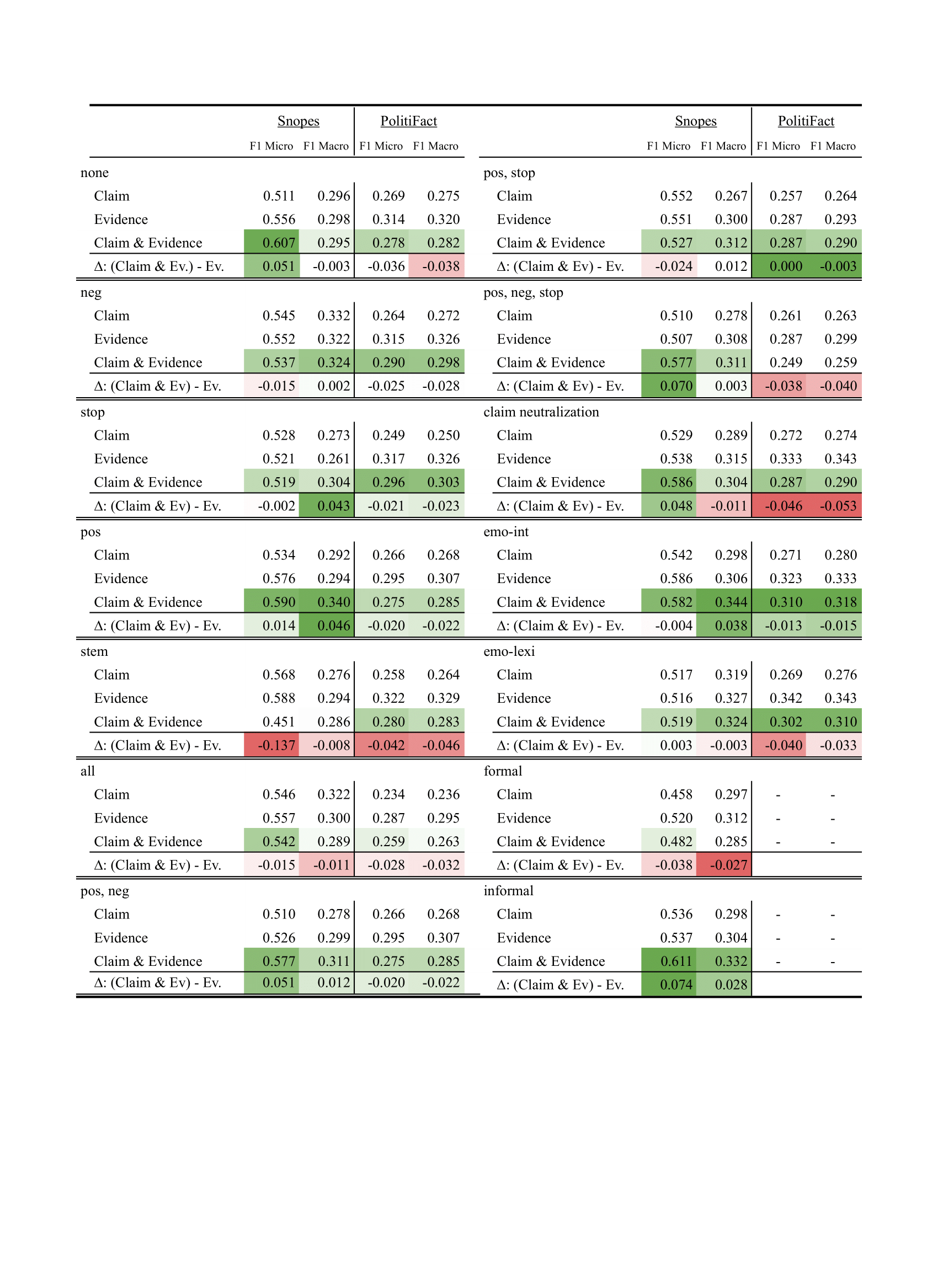}
  \label{fig:full-chart}
\end{figure*}

In Figure \ref{fig:full-chart}, we report all results for each pre-processing step.

% \clearpage 
% Entries for the entire Anthology, followed by custom entries
\bibliography{custom} %anthology
\bibliographystyle{acl_natbib}

\end{document}